\pdfoutput=1
\documentclass[11pt]{article}
\usepackage{acl2016}
\usepackage{times}
\usepackage{latexsym}
\usepackage{amsmath}
\usepackage{graphicx}
\graphicspath{ {images/} }
\usepackage{verbatim}
\usepackage{fancyvrb}
\usepackage{multirow}
\usepackage{multicol}
\usepackage{hhline}
\usepackage{url}
\usepackage{color}
\usepackage[font=small,labelfont=bf]{caption}
\usepackage{tikz}
\usepackage{standalone}
\usepackage{tikz-dependency}
\usepackage[titletoc,toc,title]{appendix}

\aclfinalcopy

\newcommand{\specialcell}[2][c]{%
  \begin{tabular}[#1]{@{}c@{}}#2\end{tabular}}

\title{Improving Hypernymy Detection\\with an Integrated Path-based and Distributional Method}

\author{Vered Shwartz ~~~ Yoav Goldberg ~~~ Ido Dagan \\
Computer Science Department \\ Bar-Ilan University \\ Ramat-Gan, Israel \\
{\normalsize \tt vered1986@gmail.com} \ {\normalsize \tt yoav.goldberg@gmail.com} \ {\normalsize \tt dagan@cs.biu.ac.il}
}

\date{}

\begin{document}

\maketitle

\begin{abstract}

Detecting hypernymy relations is a key task in NLP, which is addressed in the literature using two complementary approaches. 
Distributional methods, whose supervised variants are the current best performers, and path-based methods, which received less research attention.
We suggest an improved path-based algorithm, in which the dependency paths are encoded using a recurrent neural network, that achieves results comparable to distributional methods.
We then extend the approach to integrate both path-based and distributional signals, significantly improving upon the state-of-the-art on this task.

\end{abstract}

\section{Introduction}

Hypernymy is an important lexical-semantic relation for NLP tasks. For instance, knowing that \emph{Tom Cruise} is an \emph{actor} can help a question answering system answer the question ``which actors are involved in Scientology?''. While semantic taxonomies, like WordNet \cite{fellbaum1998wordnet}, define hypernymy relations between word types, they are limited in scope and domain. Therefore, automated methods have been developed to determine, for a given term-pair $(x,y)$, whether $y$ is an hypernym of $x$, based on their occurrences in a large corpus.

For a couple of decades, this task has been addressed by two types of approaches: distributional, and path-based. In distributional methods, the decision whether $y$ is a hypernym of $x$ is based on the distributional representations of these terms. 
Lately, with the popularity of word embeddings \cite{mikolov2013distributed}, most focus has shifted towards supervised distributional methods, in which each $(x,y)$ term-pair is represented using some combination of the terms' embedding vectors.

In contrast to distributional methods, in which the decision is based on the \emph{separate} contexts of $x$ and $y$, path-based methods base the decision on the lexico-syntactic paths connecting the \emph{joint} occurrences of $x$ and $y$ in a corpus. \newcite{hearst1992automatic} identified a small set of frequent paths that indicate hypernymy, e.g. \emph{Y such as X}. \newcite{snow2004learning} represented each $(x, y)$ term-pair as the multiset of dependency paths connecting their co-occurrences in a corpus, and trained a classifier to predict hypernymy, based on these features.

Using individual paths as features results in a huge, sparse feature space. 
While some paths are rare, they often consist of certain unimportant components. 
For instance, ``Spelt is a species of wheat'' and ``Fantasy is a genre of fiction'' yield two different paths: \emph{X be species of Y} and \emph{X be genre of Y}, while both indicating that X is-a Y. A possible solution is to generalize paths by replacing words along the path with their part-of-speech tags or with wild cards, as done in the PATTY system \cite{nakashole2012patty}. 

Overall, the state-of-the-art path-based methods perform worse than the distributional ones. 
This stems from a major limitation of path-based methods: they require that the terms of the pair occur together in the corpus, limiting the recall of these methods.
While distributional methods have no such requirement, they are usually less precise in detecting a specific semantic relation like hypernymy, and perform best on detecting broad semantic similarity between terms. Though these approaches seem complementary, there has been rather little work on integrating them \cite{mirkin2006integrating,kaji2008using}.

In this paper, we present HypeNET, an integrated path-based and distributional method for hypernymy detection. Inspired by recent progress in relation classification, we use a long short-term memory (LSTM) network \cite{hochreiter1997long} to encode dependency paths. In order to create enough training data for our network, we followed previous methodology of constructing a dataset based on knowledge resources.

We first show that our path-based approach, on its own, substantially improves performance over prior path-based methods, yielding performance comparable to state-of-the-art distributional methods. Our analysis suggests that the neural path representation enables better generalizations. While coarse-grained generalizations, such as replacing a word by its POS tag, capture mostly syntactic similarities between paths, HypeNET captures also semantic similarities.

We then show that we can easily integrate distributional signals in the network. The integration results confirm that the distributional and path-based signals indeed provide complementary information, with the combined model yielding an improvement of up to 14 $F_1$ points over each individual model.\footnote{Our code and data are available in:\\\url{https://github.com/vered1986/HypeNET}}

\section{Background}
\label{sec:background}

We introduce the two main approaches for hypernymy detection: distributional (Section \ref{sec:distributional}), and path-based (Section \ref{sec:path-based}). We then discuss the recent use of recurrent neural networks in the related task of relation classification (Section \ref{sec:relation-classification}).

\subsection{Distributional Methods}
\label{sec:distributional}

Hypernymy detection is commonly addressed using distributional methods. In these methods, the decision whether $y$ is a hypernym of $x$ is based on the distributional representations of the two terms, i.e., the contexts with which each term occurs \emph{separately} in the corpus.

Earlier methods developed unsupervised measures for hypernymy, starting with symmetric similarity measures \cite{lin1998information}, and followed by directional measures based on the distributional inclusion hypothesis \cite{weeds2003general,kotlerman2010directional}. This hypothesis states that the contexts of a hyponym are expected to be largely included in those of its hypernym. More recent work \cite{santus2014chasing,rimell2014distributional} introduce new measures, based on the assumption that the most typical linguistic contexts of a hypernym are less informative than those of its hyponyms.

More recently, the focus of the distributional approach shifted to supervised methods. In these methods, the $(x,y)$ term-pair is represented by a feature vector, and a classifier is trained on these vectors to predict hypernymy. Several methods are used to represent term-pairs as a combination of each term's embeddings vector: concatenation $\vec{x} \oplus \vec{y}$ \cite{baroni2012entailment}, difference $\vec{y} - \vec{x}$ \cite{roller2014inclusive,weeds2014learning}, and dot-product $\vec{x} \cdot \vec{y}$. Using neural word embeddings \cite{mikolov2013distributed,pennington2014glove}, these methods are easy to apply, and show good results \cite{baroni2012entailment,roller2014inclusive}.

\subsection{Path-based Methods}
\label{sec:path-based}

A different approach to detecting hypernymy between a pair of terms $(x, y)$ considers the lexico-syntactic paths that connect the \emph{joint} occurrences of $x$ and $y$ in a large corpus. Automatic acquisition of hypernyms from free text, based on such paths, was first proposed by \newcite{hearst1992automatic}, who identified a small set of lexico-syntactic paths that indicate hypernymy relations (e.g. \emph{Y such as X}, \emph{X and other Y}). 

In a later work, \newcite{snow2004learning} learned to detect hypernymy. Rather than searching for specific paths that indicate hypernymy, they represent each $(x, y)$ term-pair as the multiset of all dependency paths that connect $x$~and~$y$ in the corpus, and train a logistic regression classifier to predict whether $y$ is a hypernym of $x$, based on these paths. 

Paths that indicate hypernymy are those that were assigned high weights by the classifier. The paths identified by this method were shown to subsume those found by \newcite{hearst1992automatic}, yielding improved performance. Variations of Snow et al.'s~\shortcite{snow2004learning} method were later used in tasks such as taxonomy construction \cite{snow2006semantic,kozareva2010semi,carlson2010toward,riedel2013relation}, analogy identification \cite{turney2006similarity}, and definition extraction \cite{borg2009evolutionary,navigli2010learning}.

A major limitation in relying on lexico-syntactic paths is the sparsity of the feature space. Since similar paths may somewhat vary at the lexical level, generalizing such variations into more abstract paths can increase recall. The PATTY algorithm \cite{nakashole2012patty} applied such generalizations for the purpose of acquiring a taxonomy of term relations from free text. For each path, they added generalized versions in which a subset of words along the path were replaced by either their POS tags, their ontological types or wild-cards. This generalization increased recall while maintaining the same level of precision. 

\begin{figure}[t]
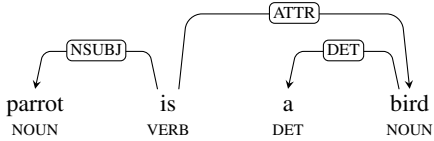
 
\centering
\small
\begin{dependency}
\begin{deptext}[column sep=3em]
    parrot \& is \& a \& bird \\
		\tiny NOUN \& \tiny VERB \& \tiny DET \& \tiny NOUN \\
  \end{deptext}
  \depedge{2}{1}{NSUBJ}
  \depedge{2}{4}{ATTR}
  \depedge{4}{3}{DET}
\end{dependency}
\caption{An example dependency tree of the sentence ``parrot is a bird'', with \emph{x=parrot} and \emph{y=bird}, represented in our notation as \scriptsize\texttt{X/NOUN/nsubj/< be/VERB/ROOT/- Y/NOUN/attr/>}.}
\label{fig:x_is_y}
\vspace*{-10pt}
\end{figure}

\let\oldttdefault=\ttdefault
\renewcommand*\ttdefault{lcmtt}

\begin{figure*}[t] 
\hspace*{-15pt}
\centering
\resizebox{1.1\linewidth}{!}{\input{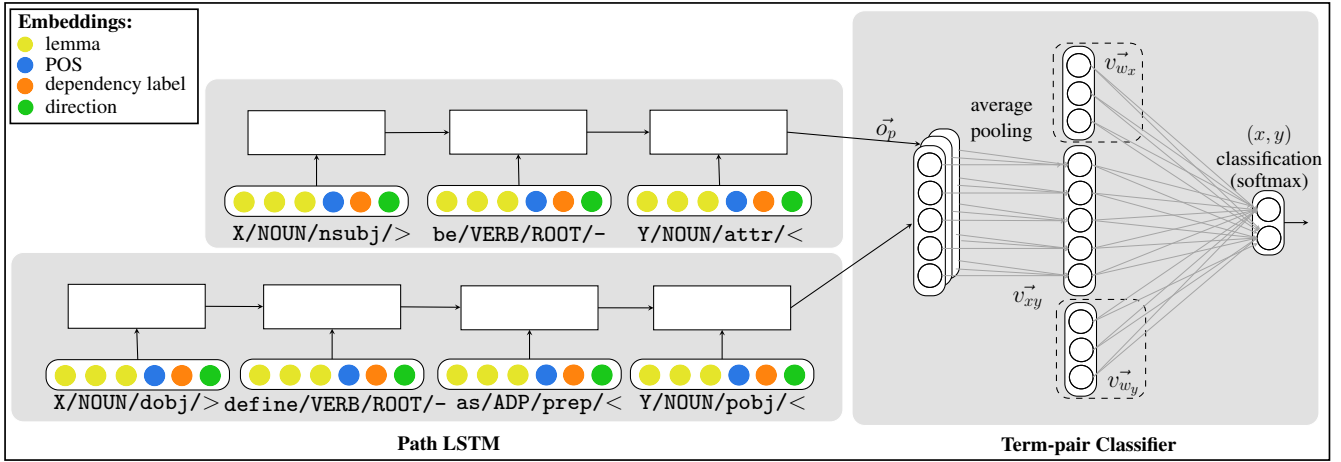}}
\caption{An illustration of term-pair classification. Each term-pair is represented by several paths. Each path is a sequence of edges, and each edge consists of four components: lemma, POS, dependency label and dependency direction. Each edge vector is fed in sequence into the LSTM, resulting in a path embedding vector $\vec{o_p}$. The averaged path vector becomes the term-pair's feature vector, used for classification. The dashed $\vec{v_{w_x}}, \vec{v_{w_y}}$ vectors refer to the integrated network described in Section~\ref{sec:combined}.}
\label{fig:lstm}
\vspace*{-5pt}
\end{figure*}

\let\ttdefault=\oldttdefault

\subsection{RNNs for Relation Classification}
\label{sec:relation-classification}

Relation classification is a related task whose goal is to classify the relation that is expressed between two target terms in a given sentence to one of predefined relation classes. To illustrate, consider the following sentence, from the SemEval-2010 relation classification task dataset \cite{hendrickx2009semeval}: ``The $[$apples$]_{e_1}$ are in the $[$basket$]_{e_2}$". Here, the relation expressed between the target entities is $Content-Container(e_1, e_2)$.

The shortest dependency paths between the target entities were shown to be informative for this task \cite{fundel2007relex}. Recently, deep learning techniques showed good performance in capturing the indicative information in such paths.

In particular, several papers show improved performance using recurrent neural networks (RNN) that process a dependency path edge-by-edge. Xu et al.~\shortcite{xu2015classifying,xu2016improved} apply a separate long short-term memory (LSTM) network to each sequence of words, POS tags, dependency labels and WordNet hypernyms along the path. A max-pooling layer on the LSTM outputs is used as the input of a network that predicts the classification. Other papers suggest incorporating additional network architectures to further improve performance \cite{nguyen2015combining,liu2015dependency}.

While relation classification and hypernymy detection are both concerned with identifying semantic relations that hold for pairs of terms, they differ in a major respect. In relation classification the relation should be expressed in the given text, while in hypernymy detection, the goal is to recognize a generic lexical-semantic relation between terms that holds in many contexts. Accordingly, in relation classification a term-pair is represented by a single dependency path, while in hypernymy detection it is represented by the multiset of all dependency paths in which they co-occur in the corpus.

\section{LSTM-based Hypernymy Detection}
\label{sec:our-method}

We present HypeNET, an LSTM-based method for hypernymy detection. We first focus on improving path representation (Section~\ref{sec:our-path-based}), and then integrate distributional signals into our network, resulting in a combined method (Section~\ref{sec:combined}).

\subsection{Path-based Network}
\label{sec:our-path-based}

Similarly to prior work, we represent each dependency path as a sequence of edges that leads from $x$ to $y$ in the dependency tree.\footnote{Like \newcite{snow2004learning}, we added for each path, additional paths containing single daughters of $x$ or $y$ not already contained in the path, referred by \newcite{snow2004learning} as ``satellite edges''. This enables including paths like \emph{Such Y as X}, in which the word ``such'' is not in the path between $x$ and $y$.} Each edge contains the lemma and part-of-speech tag of the source node, the dependency label, and the edge direction between two subsequent nodes. We denote each edge as $lemma/POS/dep/dir$. See figure~\ref{fig:x_is_y} for an illustration.

Rather than treating an entire dependency path as a single feature, we encode the sequence of edges using a long short-term memory (LSTM) network. The vectors obtained for the different paths of a given $(x,y)$ pair are pooled, and the resulting vector is used for classification. Figure~\ref{fig:lstm} depicts the overall network structure, which is described below.

\paragraph{Edge Representation}
We represent each edge by the concatenation of its components' vectors:

\vspace{-5pt}
\begin{equation*}
\vec{v_e} = [\vec{v_l}, \vec{v_{pos}}, \vec{v_{dep}}, \vec{v_{dir}}]
\end{equation*}

\noindent where $\vec{v_l}, \vec{v_{pos}}, \vec{v_{dep}}, \vec{v_{dir}}$ represent the embedding vectors of the lemma, part-of-speech, dependency label and dependency direction (along the path from $x$ to $y$), respectively. 

\paragraph{Path Representation}

For a path $p$ composed of edges $e_1,...,e_k$, the edge vectors $\vec{v_{e_1}}, ..., \vec{v_{e_k}}$ are fed in order to an LSTM encoder, resulting in a vector $\vec{o_p}$ representing the entire path $p$. The LSTM architecture is effective at capturing temporal patterns in sequences. We expect the training procedure to drive the LSTM encoder to focus on parts of the path that are more informative for the classification task while ignoring others.

\paragraph{Term-Pair Classification}
Each $(x, y)$ term-pair is represented by the multiset of lexico-syntactic paths that connected $x$ and $y$ in the corpus, denoted as $paths(x,y)$, while the supervision is given for the term pairs. We represent each $(x, y)$ term-pair as the weighted-average of its path vectors, by applying average pooling on its path vectors, as follows: 

\vspace{-10pt}
\begin{equation}
\resizebox{0.89\hsize}{!}{%
$\vec{v_{xy}} = \vec{v}_{paths(x,y)} = 
\frac{\sum_{p \in paths(x,y)} f_{p, (x,y)} \cdot \vec{o_p}}{\sum_{p \in paths(x,y)} f_{p, (x,y)}}$%
}
\label{eq:fv1}
\end{equation}

\noindent where $f_{p, (x,y)}$ is the frequency of $p$ in $paths(x,y)$. We then feed this path vector to a single-layer network that performs binary classification to decide whether $y$ is a hypernym of $x$. 

\vspace{-5pt}
\begin{equation}
c = softmax(W \cdot \vec{v_{xy}})
\label{eq:network}
\end{equation}

\noindent $c$ is a 2-dimensional vector whose components sum to 1, and we classify a pair as positive if $c[1] > 0.5$.

\paragraph{Implementation Details} To train the network, we used PyCNN.\footnote{\url{https://github.com/clab/cnn}} We minimize the cross entropy loss using gradient-based optimization, with mini-batches of size 10 and the Adam update rule \cite{kingma2014adam}. Regularization is applied by a dropout on each of the components' embeddings. We tuned the hyper-parameters (learning rate and dropout rate) on the validation set (see the appendix for the hyper-parameters values).

We initialized the lemma embeddings with the pre-trained GloVe word embeddings \cite{pennington2014glove}, trained on Wikipedia. We tried both the 50-dimensional and 100-dimensional embedding vectors and selected the ones that yield better performance on the validation set.\footnote{Higher-dimensional embeddings seem not to improve performance, while hurting the training runtime.} The other embeddings, as well as out-of-vocabulary lemmas, are initialized randomly. We update all embedding vectors during training.

\subsection{Integrated Network}
\label{sec:combined}

The network presented in Section~\ref{sec:our-path-based} classifies each $(x, y)$ term-pair based on the paths that connect $x$~and~$y$ in the corpus. Our goal was to improve upon previous path-based methods for hypernymy detection, and we show in Section~\ref{sec:results} that our network indeed outperforms them. Yet, as path-based and distributional methods are considered complementary, we present a simple way to integrate distributional features in the network, yielding improved performance.

We extended the network to take into account distributional information on each term. Inspired by the supervised distributional concatenation method \cite{baroni2012entailment}, we simply concatenate $x$ and $y$ word embeddings to the $(x, y)$ feature vector, redefining $\vec{v_{xy}}$:

\vspace{-5pt}
\begin{equation}
\vec{v_{xy}} = [\vec{v_{w_x}}, \vec{v}_{paths(x,y)}, \vec{v_{w_y}}]
\label{eq:fv2}
\end{equation}

\noindent where $\vec{v_{w_x}}$ and $\vec{v_{w_y}}$ are $x$ and $y$'s word embeddings, respectively, and $\vec{v}_{paths(x,y)}$ is the averaged path vector defined in equation~\ref{eq:fv1}. This way, each $(x, y)$ pair is represented using both the distributional features of $x$ and $y$, and their path-based features.

\section{Dataset}
\label{sec:dataset}

\subsection{Creating Instances}
\label{sec:instances}

Neural networks typically require a large amount of training data, whereas the existing hypernymy datasets, like BLESS \cite{baroni2011we}, are relatively small. 
Therefore, we followed the common methodology of creating a dataset using distant supervision from knowledge resources \cite{snow2004learning,riedel2013relation}.
Following \newcite{snow2004learning}, who constructed their dataset based on WordNet hypernymy, and aiming to create a larger dataset, we extract hypernymy relations from several resources: WordNet \cite{fellbaum1998wordnet}, DBPedia \cite{auer2007dbpedia}, Wikidata \cite{vrandevcic2012wikidata} and Yago \cite{suchanek2007yago}.

\begin{table}
\center
\small
\begin{tabular}{ | c | c | }
    \hline
     \textbf{resource} & \textbf{relations} \\ \hline
		WordNet & instance hypernym, hypernym \\ \hline
		DBPedia & type\\ \hline
		Wikidata & subclass of, instance of \\ \hline	
		Yago & subclass of \\ \hline
	\end{tabular}
	\caption{Hypernymy relations in each resource.}
	\label{tab:relations}
	\vspace{-8pt}
\end{table}

All instances in our dataset, both positive and negative, are pairs of terms that are directly related in at least one of the resources. 
These resources contain thousands of relations, some of which indicate hypernymy with varying degrees of certainty. To avoid including questionable relation types, we consider as denoting positive examples only indisputable hypernymy relations (Table~\ref{tab:relations}), which we manually selected from the set of hypernymy indicating relations in \newcite{shwartz2015learning}.

Term-pairs related by other relations (including hyponymy), are considered as negative instances. Using related rather than random term-pairs as negative instances tests our method's ability to distinguish between hypernymy and other kinds of semantic relatedness. We maintain a ratio of 1:4 positive to negative pairs in the dataset.

Like \newcite{snow2004learning}, we include only term-pairs that have joint occurrences in the corpus, requiring at least two different dependency paths for each pair.

\subsection{Random and Lexical Dataset Splits}
\label{sec:train-test-validation}

As our primary dataset, we perform standard random splitting, with 70\% train, 25\% test and 5\% validation sets. 

\begin{table}
\center
\small
\begin{tabular}{ c | c | c | c | c | }
    \hhline{~----}
    & \textbf{train} & \textbf{test} & \textbf{validation} & \textbf{all} \\ \hline
		\multicolumn{1}{|c|}{\textbf{random split}} & 49,475 & 17,670 & 3,534 & 70,679 \\ \hline
		\multicolumn{1}{|c|}{\textbf{lexical split}} & 20,335 & 6,610 & 1,350 & 28,295 \\ \hline
\end{tabular}
	\caption{The number of instances in each dataset.}
	\label{tab:dataset-size}
	\vspace{-10pt}
\end{table}

\begin{table*}[!th]
\small
\center
\begin{tabular}{ | c |}
    \hline
     \textbf{path}  \\ \hline
		\texttt{X/NOUN/dobj/> establish/VERB/ROOT/- as/ADP/prep/< Y/NOUN/pobj/<} \\ \hline
		\texttt{X/NOUN/dobj/> VERB as/ADP/prep/< Y/NOUN/pobj/<} \\ \hline
		\texttt{X/NOUN/dobj/> * as/ADP/prep/< Y/NOUN/pobj/<}  \\ \hline
		\texttt{X/NOUN/dobj/> establish/VERB/ROOT/- ADP Y/NOUN/pobj/<} \\ \hline
\end{tabular}
	\caption{Example generalizations of \emph{X was established as Y}.}
	\label{tab:gen-example}
	\vspace*{-8pt}
\end{table*}

As pointed out by \newcite{levy2015supervised}, supervised distributional lexical inference methods tend to perform ``lexical memorization'', i.e., instead of learning a relation between the two terms, they mostly learn an independent property of a single term in the pair: whether it is a ``prototypical hypernym'' or not. For instance, if the training set contains term-pairs such as \emph{(dog, animal)}, \emph{(cat, animal)}, and \emph{(cow, animal)}, all annotated as positive examples, the algorithm may learn that \emph{animal} is a prototypical hypernym, classifying any new \emph{(x, animal)} pair as positive, regardless of the relation between $x$ and \emph{animal}. \newcite{levy2015supervised} suggested to split the train and test sets such that each will contain a distinct vocabulary (``lexical split''), in order to prevent the model from overfitting by lexical memorization.

To investigate such behaviors, we present results also for a lexical split of our dataset. In this case, we split the train, test and validation sets such that each contains a distinct vocabulary. We note that this differs from \newcite{levy2015supervised}, who split only the train and the test sets, and dedicated a subset of the train for validation. We chose to deviate from \newcite{levy2015supervised} because we noticed that when the validation set contains terms from the train set, the model is rewarded for lexical memorization when tuning the hyper-parameters, consequently yielding suboptimal performance on the lexically-distinct test set. When each set has a distinct vocabulary, the hyper-parameters are tuned to avoid lexical memorization and are likely to perform better on the test set. We tried to keep roughly the same 70/25/5 ratio in our lexical split.\footnote{The lexical split discards many pairs consisting of cross-set terms.} The sizes of the two datasets are shown in Table~\ref{tab:dataset-size}.

Indeed, training a model on a lexically split dataset may result in a more general model, that can better handle pairs consisting of two unseen terms during inference. However, we argue that in the common applied scenario, the inference involves an unseen pair $(x, y)$, in which $x$ and/or $y$ have already been observed separately. Models trained on a random split may introduce the model with a term's ``prior probability'' of being a hypernym or a hyponym, and this information can be exploited beneficially at inference time.

\section{Baselines}
\label{sec:baselines}

We compare HypeNET with several state-of-the-art methods for hypernymy detection, as described in Section~\ref{sec:background}: path-based methods (Section~\ref{sec:path-based-baseline}), and distributional methods (Section~\ref{sec:distributional-baselines}). Due to different works using different datasets and corpora, we replicated the baselines rather than comparing to the reported results. 

We use the Wikipedia dump from May 2015 as the underlying corpus of all the methods, and parse it using spaCy.\footnote{\url{https://spacy.io/}}
We perform model selection on the validation set to tune the hyper-parameters of each method.\footnote{We applied grid search for a range of values, and picked the ones that yield the highest $F_1$ score on the validation set.} The best hyper-parameters are reported in the appendix.

\subsection{Path-based Methods}
\label{sec:path-based-baseline}

\paragraph{Snow} We follow the original paper, and extract all shortest paths of four edges or less between terms in a dependency tree. Like \newcite{snow2004learning}, we add paths with ``satellite edges'', i.e., single words not already contained in the dependency path, which are connected to either X or Y, allowing paths like \emph{such Y as X}. The number of distinct paths was 324,578. We apply $\chi^2$ feature selection to keep only the 100,000 most informative paths and train a logistic regression classifier. 

\paragraph{Generalization} We also compare our method to a baseline that uses generalized dependency paths. Following PATTY's approach to generalizing paths \cite{nakashole2012patty}, we replace edges with their part-of-speech tags as well as with wild cards. We generate the powerset of all possible generalizations, including the original paths. See Table~\ref{tab:gen-example} for examples. The number of features after generalization went up to 2,093,220. Similarly to the first baseline, we apply feature selection, this time keeping the 1,000,000 most informative paths, and train a logistic regression classifier over the generalized paths.\footnote{We also tried keeping the 100,000 most informative paths, but the performance was worse.}

\subsection{Distributional Methods}
\label{sec:distributional-baselines}

\begin{table*}[!ht]
\center
\small
\begin{tabular}{ c c | c | c | c || c | c | c | }
    \hhline{~~------}
		& & \multicolumn{3}{c||}{\textbf{random split}} & \multicolumn{3}{c|}{\textbf{lexical split}} \\ \hline
    \multicolumn{2}{|c|}{\textbf{method}} & \textbf{precision} & \textbf{recall} & \boldmath $F_1$ & \textbf{precision} & \textbf{recall} & \boldmath $F_1$ \\ \hline
		\multicolumn{1}{|c|}{\multirow{3}{*}{Path-based}} & \multicolumn{1}{l|}{Snow} & 0.843 & 0.452 & 0.589 & 0.760 & 0.438 & \multicolumn{1}{c|}{0.556} \\ \cline{2-8}	
		\multicolumn{1}{|l|}{} & \multicolumn{1}{l|}{Snow + Gen} & 0.852 & 0.561 & 0.676 & 0.759 & 0.530 & \multicolumn{1}{c|}{0.624} \\ \cline{2-8}
		\multicolumn{1}{|l|}{} & \multicolumn{1}{l|}{HypeNET Path-based} & 0.811 & 0.716 & 0.761 & 0.691 & \textbf{0.632} & \multicolumn{1}{c|}{0.660} \\ \hline	\hline
		\multicolumn{1}{|c|}{\multirow{2}{*}{Distributional}} &  
		\multicolumn{1}{l|}{SLQS \cite{santus2014chasing}} & 0.491 & 0.737 & 0.589 & 0.375 & 0.610 & \multicolumn{1}{c|}{0.464} \\ \cline{2-8}
		\multicolumn{1}{|l|}{} & \multicolumn{1}{l|}{Best supervised (concatenation)} & 0.901 & 0.637 & 0.746 & 0.754 & 0.551 & \multicolumn{1}{c|}{0.637} \\ \hline \hline
		\multicolumn{1}{|c|}{\multirow{1}{*}{Combined}} & 
		\multicolumn{1}{l|}{HypeNET Integrated} & \textbf{0.913} & \textbf{0.890} & \textbf{0.901} & \textbf{0.809} & 0.617 & \multicolumn{1}{c|}{\textbf{0.700}} \\ \hline 
	\end{tabular}
	\caption{Performance scores of our method compared to the path-based baselines and the state-of-the-art distributional methods for hypernymy detection, on both variations of the dataset -- with lexical and random split to train / test / validation.}
	\label{tab:results}
	\vspace*{-10pt}
\end{table*}

\paragraph{Unsupervised} SLQS \cite{santus2014chasing} is an entropy-based measure for hypernymy detection, reported to outperform previous state-of-the-art unsupervised methods \cite{weeds2003general,kotlerman2010directional}. The original paper was evaluated on the BLESS dataset \cite{baroni2011we}, which consists of mostly frequent words. Applying the vanilla settings of SLQS on our dataset, that contains also rare terms, resulted in low performance. 
Therefore, we received assistance from Enrico Santus, who kindly provided the results of SLQS on our dataset after tuning the system as follows.

The validation set was used to tune the threshold for classifying a pair as positive, as well as the maximum number of each term's most associated contexts ($N$). In contrast to the original paper, in which the number of each term's contexts is fixed to $N$, in this adaptation it was set to the minimum between the number of contexts with LMI score above zero and $N$. In addition, the SLQS scores were not multiplied by the cosine similarity scores between terms, and terms were lemmatized prior to computing the SLQS scores, significantly improving recall.

As our results suggest, while this method is state-of-the-art for unsupervised hypernymy detection, it is basically designed for classifying specificity level of related terms, rather than hypernymy in particular.

\paragraph{Supervised} To represent term-pairs with distributional features, we tried several state-of-the-art methods: concatenation $\vec{x} \oplus \vec{y}$ \cite{baroni2012entailment}, difference $\vec{y} - \vec{x}$ \cite{roller2014inclusive,weeds2014learning}, and dot-product $\vec{x} \cdot \vec{y}$. We downloaded several pre-trained embeddings \cite{mikolov2013distributed,pennington2014glove} of different sizes, and trained a number of classifiers: logistic regression, SVM, and SVM with RBF kernel, which was reported by \newcite{levy2015supervised} to perform best in this setting. We perform model selection on the validation set to select the best vectors, method and regularization factor (see the appendix).

\section{Results}
\label{sec:results}

Table~\ref{tab:results} displays performance scores of HypeNET and the baselines. \emph{HypeNET Path-based} is our path-based recurrent neural network model (Section~\ref{sec:our-path-based}) and \emph{HypeNET Integrated} is our combined method (Section~\ref{sec:combined}). Comparing the path-based methods shows that generalizing paths improves recall while maintaining similar levels of precision, reassessing the behavior found in \newcite{nakashole2012patty}. \emph{HypeNET Path-based} outperforms both path-based baselines by a significant improvement in recall and with slightly lower precision. The recall boost is due to better path generalization, as demonstrated in Section~\ref{sec:qualitative}.

Regarding distributional methods, the unsupervised SLQS baseline performed slightly worse on our dataset. 
The low precision stems from its inability to distinguish between hypernyms and meronyms, which are common in our dataset, causing many false positive pairs such as \emph{(zabrze, poland)} and \emph{(kibbutz, israel)}. We sampled 50 false positive pairs of each dataset split, and found that 38\% of the false positive pairs in the random split and 48\% of those in the lexical split were holonym-meronym pairs. 

In accordance with previously reported results, the supervised embedding-based method is the best performing baseline on our dataset as well. \emph{HypeNET Path-based} performs slightly better, achieving state-of-the-art results. Adding distributional features to our method shows that these two approaches are indeed complementary. On both dataset splits, the performance differences between \emph{HypeNET Integrated} and \emph{HypeNET Path-based}, as well as the supervised distributional method, are substantial, and statistically significant with p-value of 1\% (paired t-test).

We also reassess that indeed supervised distributional methods perform worse on a lexical split \cite{levy2015supervised}. We further observe a similar reduction when using HypeNET, which is not a result of lexical memorization, but rather stems from over-generalization (Section~\ref{sec:qualitative}).

\begin{table*}[!ht]
\hspace*{-20pt}
\small
\begin{tabular}{ | c | c | c | }
    \hline
    \textbf{method} & \textbf{path} & \textbf{example text} \\ \hline
		\multirow{3}{*}{Snow}
		& \specialcell{\texttt{X/NOUN/nsubj/> be/VERB/ROOT/- Y/NOUN/attr/<} \\ \texttt{direct/VERB/acl/>}} & 
		\multicolumn{1}{c|}{\specialcell{\textcolor{blue}{\emph{Eyeball}} is a 1975 Italian-Spanish\\\textcolor{red}{\emph{film}} directed by Umberto Lenzi}} \\ \cline{2-3}
		& \specialcell{\texttt{X/NOUN/nsubj/> be/VERB/ROOT/- Y/NOUN/attr/<} \\ \texttt{publish/VERB/acl/>}} & 
		\multicolumn{1}{c|}{\specialcell{\textcolor{blue}{\emph{Allure}} is a U.S. women's beauty\\\textcolor{red}{\emph{magazine}} published monthly}} \\ \hline
		\multirow{2}{*}{\specialcell{Snow +\\Gen}}
		& \specialcell{\texttt{X/NOUN/compound/> NOUN$^*$ be/VERB/ROOT/-} \\ \texttt{Y/NOUN/attr/< base/VERB/acl/>}}
		& \multicolumn{1}{c|}{\
		\specialcell{\textcolor{blue}{\emph{Calico}} Light Weapons Inc. (CLWS) is an\\American privately held manufacturing\\ \textcolor{red}{\emph{company}} based in Cornelius, Oregon}} \\ \cline{2-3}
		& \specialcell{\specialcell{\texttt{X/NOUN/compound/> NOUN Y/NOUN/compound/<}}} 
		& \multicolumn{1}{c|}{\specialcell{\textcolor{blue}{\emph{Weston}} \textcolor{red}{\emph{Town}} Council}} \\ \hline
		\multirow{6}{*}{\specialcell{HypeNET\\Integrated}} 
		& \specialcell{\texttt{X/NOUN/nsubj/> be/VERB/ROOT/- Y/NOUN/attr/<} \\ \texttt{(release|direct|produce|write)/VERB/acl/>}} & 
		\multicolumn{1}{c|}{\specialcell{\textcolor{blue}{\emph{Blinky}} is a 1923 American comedy\\ \textcolor{red}{\emph{film}} directed by Edward Sedgwick}} \\ \cline{2-3}
		& \specialcell{\texttt{X/NOUN/compound/>} \\ \texttt{(association|co.|company|corporation|foundation} \\ \texttt{|group|inc.|international|limited|ltd.)/NOUN/nsubj/>} \\ 
		\texttt{be/VERB/ROOT/- Y/NOUN/attr/<} \\ \texttt{((create|found|headquarter|own|specialize)/VERB/acl/>)?}} &
		\multicolumn{1}{c|}{\textcolor{blue}{\emph{Retalix}} Ltd. is a software \textcolor{red}{\emph{company}}} \\ 
		\hline
	\end{tabular}
	\caption{Examples of indicative paths learned by each method, with corresponding true positive term-pairs from the random split test set. Hypernyms are marked red and hyponyms are marked blue.}
	\label{tab:prominent-paths}
	\vspace*{-15pt}
\end{table*}

\section{Analysis}
\label{sec:analysis}

\subsection{Qualitative Analysis of Learned Paths}
\label{sec:qualitative}

We analyze HypeNET's ability to generalize over path structures, by comparing prominent indicative paths which were learned by each of the path-based methods. We do so by finding high-scoring paths that contributed to the classification of true-positive pairs in the dataset. In the path-based baselines, these are the highest-weighted features as learned by the logistic regression classifier. In the LSTM-based method, it is less straightforward to identify the most indicative paths. We assess the contribution of a certain path $p$ to classification by regarding it as the only path that appeared for the term-pair, and compute its \textsc{true} label score from the class distribution: $softmax(W \cdot \vec{v_{xy}})[1]$, setting $\vec{v_{xy}} = [\vec{0}, \vec{o_p}, \vec{0}]$. 

A notable pattern is that Snow's method learns specific paths, like \emph{X is Y from} (e.g. \emph{\textbf{Megadeth} is an American thrash metal \textbf{band} from Los Angeles}). While Snow's method can only rely on verbatim paths, limiting its recall, the generalized version of Snow often makes coarse generalizations, such as \emph{X VERB Y from}. Clearly, such a path is too general, and almost any verb assigned to it results in a non-indicative path (e.g. \emph{X take Y from}). Efforts by the learning method to avoid such generalization, again, lower the recall. HypeNET provides a better midpoint, making fine-grained generalizations by learning additional semantically similar paths such as \emph{X become Y from} and \emph{X remain Y from}. See table~\ref{tab:prominent-paths} for additional example paths which illustrate these behaviors.

We also noticed that while on the random split our model learns a range of specific paths such as \emph{X is Y published} (learned for e.g. \emph{Y=magazine}) and \emph{X is Y produced} (\emph{Y=film}), in the lexical split it only learns the general \emph{X is Y} path for these relations. We note that \emph{X is Y} is a rather ``noisy'' path, which may occur in ad-hoc contexts without indicating generic hypernymy relations (e.g. \emph{\textbf{chocolate} is a big \textbf{problem}} in the context of children's health). 
While such a model may identify hypernymy relations between unseen terms, based on general paths, it is prone to over-generalization, hurting its performance, as seen in Table~\ref{tab:results}. As discussed in \S~\ref{sec:train-test-validation}, we suspect that this scenario, in which both terms are unseen, is usually not common enough to justify this limiting training setup.

\subsection{Error Analysis}
\label{sec:error_analysis}

\begin{table}[t]
\center
\small
\begin{tabular}{|c|c|}
\hline
\textbf{Relation} & \textbf{\%} \\
\hline
synonymy & $21.37\%$ \\
hyponymy & $29.45\%$ \\
holonymy / meronymy & $9.36\%$ \\
hypernymy-like relations & $21.03\%$ \\
other relations & $18.77\%$ \\
\hline
\end{tabular}
\caption{Distribution of relations holding between each pair of terms in the resources among false positive pairs.}
\label{tab:false-positives}
\vspace*{-10pt}
\end{table}

\paragraph{False Positives}

We categorized the false positive pairs on the random split according to the relation holding between each pair of terms in the resources used to construct the dataset. 
We grouped several semantic relations from different resources to broad categories, e.g. \emph{synonym} includes also \emph{alias} and \emph{Wikipedia redirection}.
Table~\ref{tab:false-positives} displays the distribution of semantic relations among false positive pairs. 

More than 20\% of the errors stem from confusing synonymy with hypernymy, which are known to be difficult to distinguish. 

An additional 30\% of the term-pairs are reversed hypernym-hyponym pairs ($y$ is a hyponym of $x$). Examining a sample of these pairs suggests that they are usually near-synonyms, i.e., it is not that clear whether one term is truely more general than the other or not. For instance, \emph{fiction} is annotated in WordNet as a hypernym of \emph{story}, while our method classified \emph{fiction} as its hyponym. 

A possible future research direction might be to quite simply extend our network to classify term-pairs simultaneously to multiple semantic relations, as in \newcite{PavlickEtAl-2015:ACL:Semantics}. Such a multiclass model can hopefully better distinguish between these similar semantic relations.

Another notable category is hypernymy-like relations: these are other relations in the resources that could also be considered as hypernymy, but were annotated as negative due to our restrictive selection of only indisputable hypernymy relations from the resources (see Section~\ref{sec:instances}). These include instances like \emph{(Goethe, occupation, novelist)} and \emph{(Homo, subdivisionRanks, species)}. 

Lastly, other errors made by the model often correspond to term-pairs that co-occur very few times in the corpus, e.g. $xebec$, a studio producing Anime, was falsely classified as a hyponym of \emph{anime}.

\begin{table}[t]
\center
\small
\begin{tabular}{|c|c|c|}
\hline
& \textbf{Error Type} & \textbf{\%} \\
\hline
1 & low statistics & $80\%$ \\
2 & infrequent term & $36\%$ \\
3 & rare hyponym sense & $16\%$ \\
4 & annotation error & $8\%$ \\
\hline
\end{tabular}
\caption{(Overlapping) categories of false negative pairs:
(1) $x$ and $y$ co-occurred less than 25 times (average co-occurrences for true positive pairs is 99.7).
(2) Either $x$ or $y$ is infrequent.
(3) The hypernymy relation holds for a rare sense of $x$.
(4) $(x,y)$ was incorrectly annotated as positive. }
\label{tab:false-negatives}
\vspace*{-10pt}
\end{table}

\paragraph{False Negatives}
We sampled 50 term-pairs that were falsely annotated as negative, and analyzed the major (overlapping) types of errors (Table~\ref{tab:false-negatives}).

Most of these pairs had only few co-occurrences in the corpus. This is often either due to infrequent terms (e.g. \emph{cbc.ca}), or a rare sense of $x$ in which the hypernymy relation holds (e.g. \emph{(night, play)} holding for ``Night'', a dramatic sketch by Harold Pinter). Such a term-pair may have too few hypernymy-indicating paths, leading to classifying it as negative.

\section{Conclusion}

We presented HypeNET: a neural-networks-based method for hypernymy detection. First, we focused on improving path representation using LSTM, resulting in a path-based model that performs significantly better than prior path-based methods, and matches the performance of the previously superior distributional methods. In particular, we demonstrated that the increase in recall is a result of generalizing semantically-similar paths, in contrast to prior methods, which either make no generalizations or over-generalize paths. 

We then extended our network by integrating distributional signals, yielding an improvement of additional 14 $F_1$ points, and demonstrating that the path-based and the distributional approaches are indeed complementary. 

Finally, our architecture seems straightforwardly applicable for multi-class classification, which, in future work, could be used to classify term-pairs to multiple semantic relations.

\section*{Acknowledgments}
We would like to thank Omer Levy for his involvement and assistance in the early stage of this project and Enrico Santus for helping us by computing the results of SLQS \cite{santus2014chasing} on our dataset.

This work was partially supported by an Intel ICRI-CI grant, the Israel Science Foundation grant 880/12, and the German Research Foundation through the German-Israeli Project Cooperation (DIP, grant DA 1600/1-1).

\bibliography{hypernymy-detection}
\bibliographystyle{acl2016}

\begin{appendices}

\section{Best Hyper-parameters}
\label{sec:hyper-params}

Table~\ref{tab:features} displays the chosen hyper-parameters of each method, yielding the highest $F_1$ score on the validation set.

\begin{table}[!h]
\center
\scriptsize
\begin{tabular}{ | c | c | c | }
    \hline
		& \textbf{method} & \textbf{values} \\ \hline
		\multirow{10}{*}{\rotatebox[origin=c]{90}{\textbf{random split}}} & Snow & regularization: $L_2$ \\ \cline{2-3}
		& Snow + Gen & regularization: $L_1$ \\ \cline{2-3}
		& \specialcell{LSTM} & 
		\specialcell{embeddings: GloVe-100-Wiki\\learning rate: $\alpha=0.001$\\dropout: $d=0.5$} \\ \cline{2-3}
		& \specialcell{SLQS} & \specialcell{N=100, threshold = 0.000464} \\ \cline{2-3}
		& \specialcell{Best\\Supervised} & \specialcell{method: concatenation, classifier: SVM\\embeddings: GloVe-300-Wiki} \\ \cline{2-3}
		& \specialcell{LSTM-\\Integrated} & 
		\specialcell{embeddings: GloVe-50-Wiki\\learning rate: $\alpha=0.001$\\word dropout: $d=0.3$} \\ \hline
		\multirow{10}{*}{\rotatebox[origin=c]{90}{\textbf{lexical split}}} & Snow & regularization: $L_2$ \\ \cline{2-3}
		& Snow + Gen & regularization: $L_2$ \\ \cline{2-3}
		& \specialcell{LSTM} & 
		\specialcell{embeddings: GloVe-50-Wiki\\learning rate: $\alpha=0.001$\\dropout: $d=0.5$} \\ \cline{2-3}
		& \specialcell{SLQS} & \specialcell{N=100, threshold = 0.007629} \\ \cline{2-3}	
		& \specialcell{Best\\Supervised} & \specialcell{method: concatenation, classifier: SVM\\embeddings: GloVe-100-Wikipedia} \\ \cline{2-3}
		& \specialcell{LSTM-\\Integrated} & 
		\specialcell{embeddings: GloVe-50-Wiki\\learning rate: $\alpha=0.001$\\word dropout: $d=0.3$} \\ \hline
	\end{tabular}
	\caption{The best hyper-parameters in every model.}
	\label{tab:features}
	\vspace*{-15pt}
\end{table}

\end{appendices}

\end{document}